\documentclass[11pt]{article}

\usepackage[preprint]{acl}

\usepackage{times}
\usepackage{latexsym}
\usepackage{booktabs}
\usepackage{amsmath, amssymb}
\usepackage{booktabs}
\usepackage{multirow}
\usepackage{graphicx}
\usepackage{natbib}
\usepackage{hyperref}
\usepackage{xcolor}
\usepackage{multirow} 

\usepackage[T1]{fontenc}

\usepackage[utf8]{inputenc}

\usepackage{microtype}

\usepackage{inconsolata}

\usepackage{graphicx}

\title{Where does Absolute Position come from in decoder-only Transformers?}

\author{
 \textbf{Valeria Ruscio\textsuperscript{2,1}},
 \textbf{Umberto Nanni\textsuperscript{1}},
 \textbf{Fabrizio Silvestri\textsuperscript{1}}
\\
\\
\\
 \textsuperscript{1}Sapienza University of Rome,
 \textsuperscript{2}Intuition Machines
\\
 \small{
   \textbf{} \href{mailto:ruscio.valeria@gmail.com}{ruscio.valeria@gmail.com}
 }
}

\begin{document}

\maketitle

\begin{abstract}
RoPE-trained transformers distinguish absolute position in their
attention patterns, even though RoPE encodes only relative offsets in
the inner product. We trace this leakage to two architectural
components, The causal mask is responsible for the first: its per-query softmax
denominator depends on the absolute query position by construction.
The residual stream supplies the second. Under causal attention the
activation at position $0$ attends only to itself and runs as a closed
dynamical system from the embedding of the token at that position;
downstream attention reads this trajectory through sink-reading heads.
Both components appear in all three architectures we study, in
architecturally specific balance: NTK scaling suppresses the
residual-stream component, sliding-window attention allows it to
accumulate with depth, and standard RoPE sits between. Replacing the
\texttt{BOS} embedding before the forward pass removes $40\%$ of the
residual-stream component at early queries. Attention sinks are
token-anchored stabilizers that pass forward a deterministic
fingerprint of the token at position $0$, constant across inputs when
that token is the auto-prepended \texttt{BOS} and varying with it
otherwise.
\end{abstract}

\section{Introduction}

Decoder-only transformers trained with rotary position embeddings
\citep{su2024roformer} compute attention through inner products that,
by construction, depend on the difference $i - j$ of the two positions.
A model trained under this design should be unable to distinguish a
token at position $50$ from the same token at position $100$ except
through its relations to other tokens. RoPE-trained models make this
distinction readily; their attention patterns carry traces of absolute
position not predictable from relative offset alone
\citep{kazemnejad2023impact, ruscio2025beyond}. The relative encoding
leaks though: our concern in this paper is where the leakage comes from and
by what mechanism it arrives at the attention computations that
consume it.

A natural first guess identifies the attention sink as the locus.
Sinks are the token that marks the start of the sequence that absorb a large share of attention from every query \citep{xiao2024efficient, gu2025attention}, they are visited globally, appear early in the network, and
from the outside have the appearance of aggregators. We find that
sinks carry no decodable content. The positions with the highest
attention magnitude have topic-probe accuracy at chance; the positions
decodable above chance have attention magnitudes three orders of
magnitude smaller. The sink is a token-anchored feature: moving the
\texttt{BOS} token from position $0$ to position $8$ moves the sink
with it, and replacing \texttt{BOS} with a random uncommon token
suppresses the sink entirely. Sink-reading heads pass the identity of
the sink token to downstream attention: in models that auto-prepend
\texttt{BOS} this signal is constant across inputs, in Qwen it varies
with whatever token sits at position $0$.

The leakage has two architectural sources. The causal mask supplies
one: its per-query softmax denominator depends on the absolute query
position by construction, regardless of what the logits encode. The
residual stream supplies the other: position-varying activations flow
into queries and keys, and RoPE rotation does not fully obscure them.
We isolate the two components through ablations of the attention
computation, and find both present in all three architectures we study,
in architecturally specific balance.
The residual-stream component itself traces to a single mechanism.
Under causal attention, position $0$ attends only to itself; the
activation at position $0$ at every layer is a deterministic function
of the embedding of the token there. 


\section{Related Work}
\label{sec:related}
 
\paragraph{Attention sinks}
\citet{xiao2024efficient} identified attention sinks as a stable feature
of decoder-only transformers and exploited them in StreamingLLM to
maintain quality under context truncation. \citet{gu2025attention}
characterized the training dynamics under which sinks emerge.
\citet{ruscio2026you} analyzed sinks as a geometric reference frame
for downstream attention. Whether sinks function as information
aggregators or as content-free stabilizers has remained ambiguous in
the literature; our results support the latter reading and identify
token-anchoring rather than position-anchoring as the empirical
signature.
 
\paragraph{Positional encodings}
RoPE \citep{su2024roformer} encodes position through relative offsets
in the inner product, and has become the dominant choice in open-source
language models. The design property of
Equation~\ref{eq:rope-relative} has nevertheless been shown to leak:
\citet{kazemnejad2023impact} demonstrated that the choice of positional
encoding affects length-generalization in ways the relative-only design
does not predict, and \citet{ruscio2025beyond} documented
absolute-position structure in attention patterns of RoPE-trained
models. ALiBi \citep{press2021train} and the NTK-/YaRN-style
extensions \citep{peng2024yarn} modify position encoding to address
length-generalization specifically. Our work locates the leakage
downstream of the inner product, in the softmax normalization rather
than in RoPE itself.
 
\paragraph{Causal masking and mechanistic interpretation}
\citet{haviv-etal-2023-understanding} showed that decoder-only transformers
trained without explicit positional encodings can still solve
sequence tasks, attributing the residual position signal to the
asymmetry the causal mask introduces. Our work identifies the
mechanism more precisely: the position-dependent cardinality of the
per-query softmax denominator, isolated by the bidirectional
ablation. On the interpretability side, the targeted-ablation
literature has localized many model behaviors to specific heads,
circuits, or low-rank directions
\citep{olsson2022context, wang2023interpretability}. The
absolute-position structure we study resists this kind of
localization across two separate intervention families, consistent
with a mechanism that lives in the normalization rather than in any
weight.

\section{Background and the leakage metric}
\label{sec:mech}

RoPE applies a position-dependent rotation to queries and keys without
modifying the residual stream from which they are derived. For a head
with head dimension $d_h$ and unrotated query and key $q_m, k_n \in
\mathbb{R}^{d_h}$, RoPE multiplies each by a block-diagonal orthogonal
matrix
\begin{equation}
R(m) = \operatorname{diag}\!\bigl(R_2(m \theta_0),\, R_2(m \theta_1),\, \ldots,\, R_2(m \theta_{d_h/2 - 1})\bigr),
\label{eq:rope-rotation}
\end{equation}
with $R_2(\phi)$ the standard $2 \times 2$ rotation and $\theta_i =
\mathrm{base}^{-2 i / d_h}$. Because $R(m)^\top R(n) = R(n - m)$, the
rotated inner product
\begin{equation}
\bigl\langle R(m)\, q_m,\; R(n)\, k_n \bigr\rangle \;=\; q_m^\top R(n - m)\, k_n
\label{eq:rope-relative}
\end{equation}
depends on the two absolute positions only through their difference.
This is the design property of RoPE.

The post-softmax weight is a function of the entire row of logits and
the attention mask. Under causal masking,
\begin{equation}
a_{ij} \;=\; \frac{\exp(\ell_{ij})}{\sum_{k=0}^{i} \exp(\ell_{ik})}, \qquad j \leq i,
\label{eq:causal-softmax}
\end{equation}
with denominator $Z_i$ summing over $i + 1$ terms. The cardinality of
$Z_i$ depends on the absolute query position, and the post-softmax
weight inherits a dependence on $i$ not present in the logits.
Bidirectional masking sums over $T$ terms regardless of the query.
Equation~\ref{eq:causal-softmax} therefore supplies absolute-position
structure to attention weights independent of any property of the
residual stream or the rotation schedule. The residual stream supplies
a second contribution: if $x_m^{(L)}$ has acquired through previous
layer writes a component that varies with $m$, that component flows
into $q_m = W_Q^{(L,h)} x_m^{(L)}$ and into every attention logit the
head computes.

A naive linear-in-offset model of absolute-position leakage has two
problems. The predictors $i$, $j$, and $i - j$ are linearly dependent.
RoPE's contribution to the inner product is a sum of cosines at the
head's frequency-dependent angles, so a linear baseline underfits and
attributes relative-offset variance to absolute position. We use a
saturated relative-offset baseline. The baseline predicts $\ell_{ij}
= f(i - j) + \varepsilon$ with $f$ fit as one free parameter per
distinct value of $i - j$. The full model adds mean-centered $i$ and
$j$ as continuous predictors. Absolute-position leakage is
\begin{equation}
\Delta R^2(L, h) \;=\; R^2_{\text{full}}(L, h) - R^2_{\text{base}}(L, h).
\label{eq:dr2}
\end{equation}
A head with $\Delta R^2 > 0$ has access to absolute-position
information that did not arrive through Equation~\ref{eq:rope-relative}.

The two architectural sources of $\Delta R^2$ are operationally
separable. Under bidirectional attention the cardinality of the
softmax denominator is constant across queries and the causal-mask
contribution vanishes; whatever $\Delta R^2$ survives the bidirectional
ablation comes from the residual-stream component alone. We therefore
write
\begin{equation} \label{eq:decomposition}
\begin{split}
\Delta R^2_{\text{baseline}} \;=\; \Delta R^2_{\text{causal-mask}} + \Delta R^2_{\text{residual}} \\ \quad
\Delta R^2_{\text{residual}} \;=\; \Delta R^2_{\text{bidirectional}}
\end{split}
\end{equation}
with the causal-mask share given by $1 - \Delta R^2_{\text{bidirectional}}
/ \Delta R^2_{\text{baseline}}$\footnote{Equation~\ref{eq:decomposition} treats $\Delta R^2_{\text{baseline}}$
as the sum of two components and identifies
$\Delta R^2_{\text{bidirectional}}$ with the residual-stream component.
The additivity is an assumption, under
the design property of Equation~\ref{eq:rope-relative}, the two
components have orthogonal origins: the causal-mask contribution
arises from the $i$-dependent cardinality of $Z_i$ regardless of what
$f$ is, and the residual-stream contribution arises from departures of
$\ell_{ij}$ from the relative-only form. Bidirectional ablation
changes the cardinality of $Z_i$ but leaves the unrotated query/key
structure intact at first order, so $\Delta R^2_{\text{bidirec}}$
measures the residual-stream contribution under the same
residual-stream content the causal forward pass produces. The
assumption can fail if removing the causal mask materially changes the
activations the network computes on the same input through, for
example, layer-norm feedback; we do not retrain, so the weights are
fixed, but the forward pass is different. We treat additivity as
approximate and check it indirectly: identity-RoPE and scrambled-RoPE
produce nearly identical ratios in every model
(Table~\ref{tab:mechanism}), as expected if the two RoPE-disabling
interventions act on a single residual-stream contribution rather than
on a quantity that interacts with the causal-mask contribution.}.

\paragraph{The causal-mask component}
Under the design property of Equation~\ref{eq:rope-relative}, attention
logits depend only on relative offset: $\ell_{ij} = f(i - j)$ for some
function $f$. Substituting into Equation~\ref{eq:causal-softmax} gives
\begin{equation}
a_{ij} \;=\; \frac{\exp\bigl(f(i - j)\bigr)}{Z_i},
\qquad
Z_i \;=\; \sum_{d=0}^{i} \exp\bigl(f(d)\bigr).
\label{eq:partial-sum}
\end{equation}
The numerator depends on $i$ and $j$ only through their difference; the
denominator is the partial sum of $\exp(f(d))$ up to $d = i$, which
depends explicitly on $i$. If $\sum_{d=0}^{\infty} \exp(f(d))$
converges, $Z_i$ approaches its limit at a rate set by the tail of $f$
and the position dependence of $a_{ij}$ attenuates as $i$ grows. The
bidirectional ablation replaces the upper limit $i$ in $Z_i$ with the
constant $T - 1$ for every query.

\paragraph{The position-0 trajectory}
Under causal attention, position $0$ attends only to itself. The
per-layer update at position $0$ reduces to
\begin{equation}
x_0^{(\ell + 1)} \;=\; \mathrm{FFN}_\ell\!\bigl(x_0^{(\ell)} + W_O^{(\ell)} W_V^{(\ell)} x_0^{(\ell)}\bigr),
\end{equation}
a map that depends on $x_0^{(\ell)}$ alone. The trajectory $e_0 \to
x_0^{(0)} \to x_0^{(1)} \to \cdots$ is a closed dynamical system from
the embedding $e_0$. Two consequences follow. The activation
$x_0^{(\ell)}$ at every layer is a fixed function of $e_0$ across
prompts. Any downstream head that reads from position $0$ reads a
deterministic function of the token at position $0$. Section
\ref{sec:results-bosfp} confirms both empirically.

\section{Methodology}
\label{sec:methodology}

We analyze Llama-3.2-1B \citep{grattafiori2024llama} ($L = 16$, $H = 32$,
standard RoPE, auto-prepends \texttt{BOS}), Qwen-2.5-3B
\citep{hui2024qwen2} ($L = 36$, $H = 16$, NTK-scaled RoPE, no
\texttt{BOS}), and Mistral-7B-v0.3 \citep{DBLP:journals/corr/abs-2310-06825} ($L = 32$,
$H = 32$, sliding-window). The scaling analysis additionally uses
Llama-3.2-3B and Llama-3.1-8B. The mechanism-ablation,
refined-binning, layer-wise, and position-$0$ analyses use Llama-3.2-3B
in place of 1B. Corpus: $P = 64$ Wikipedia chunks balanced across
four topics, tokenized to $T = 256$. The length-scaling analysis runs
the same corpus at sixteen lengths between $56$ and $768$ tokens.

At each $(\ell, k)$ we compute the sink magnitude $\mathcal{S}(\ell, k)
= \max_{h} \frac{1}{P (T - k)} \sum_{p, q \geq k} a_{q,
k}^{(\ell, h, p)}$ and the topic decodability $\mathcal{A}(\ell, k)$,
the mean five-fold cross-validated accuracy of a logistic regression
on $x_k^{(\ell, p)}$ after PCA reduction to $64$ components; chance is
$0.25$. For $\Delta R^2(\ell, h)$ we sample $K = 4{,}000$ stratified
causal pairs, fit both regressions in double precision, and implement
the saturated baseline as one-hot indicators on distinct $i - j$.
Significance is assessed against a permutation null with $B = 500$
resamples per head, with Benjamini-Hochberg FDR at $\alpha = 0.05$.

\paragraph{Sink-variant and cosine analyses}
Three corpus variants dissociate position- from token-anchoring of the
sink: \emph{shifted} moves \texttt{BOS} to position $8$, \emph{random}
replaces the token at position $0$ with an uncommon token from the
vocabulary frequency range $[10^3, 2 \times 10^4]$. We measure cosine
consistency $\mathcal{C}(\ell, h)$ of each head's $W_O$ output at
$k^\star = 64$ across prompts, ranked by sink-reading strength
$\rho(\ell, h) = \frac{1}{P} \sum_p a_{k^\star, 0}^{(\ell, h, p)}$.

\paragraph{Qwen sink-token-identity probe}
For the top twenty Qwen sink-reading heads ranked by $\rho$, we probe
whether the head's output at $k^\star$ carries the identity of the
token at position $0$. We sample $300$ prompts with the first token
uniformly drawn from the same vocabulary frequency range and the
remaining tokens identical across prompts. We project each head's
$W_O$ output to its top principal component and report
$R^2_{\text{tok}}$ of the projection onto a one-hot encoding of the
sink-token identity, together with a random-permutation null
$R^2_{\text{random}}$. We retain heads with PC1 explained-variance
fraction above $10\%$.

\paragraph{Mechanism ablations}
\emph{Identity-RoPE} replaces $R(m)$ in Equation~\ref{eq:rope-rotation}
with the identity. \emph{Scrambled-RoPE} applies a fixed random
permutation to the assignment of frequency bands $\theta_i$ to
query/key dimensions, preserving rotation magnitudes while destroying
the relative-offset property of Equation~\ref{eq:rope-relative}.
\emph{Bidirectional} removes the causal mask. Each is applied at every
layer and head; we recompute $\Delta R^2$ on the same stratified pairs
as the baseline.

\paragraph{Refined per-query-bin and layer-wise analyses}
We aggregate $\Delta R^2$ contributions into seven bins of the query
position $i$: $[0, 5)$, $[5, 12)$, $[12, 25)$, $[25, 50)$, $[50,
100)$, $[100, 200)$, $[200, 256)$, and define the causal-mask share
within each bin via Equation~\ref{eq:decomposition}. The layer-wise
ratio $r_\ell = \mathrm{mean}_h \Delta R^2_{\text{bidir}}(\ell)
/ \mathrm{mean}_h \Delta R^2_{\text{base}}(\ell)$ measures the
residual-stream share of leakage at layer $\ell$.

\paragraph{Position-0 interventions}
To attribute the residual-stream component to specific structures in
the position-$0$ trajectory, we run three interventions on Llama-3.2-3B
under bidirectional attention and report first-bin $[0, 5)$ mean
$\Delta R^2$. \emph{bi-replace-embedding} replaces the \texttt{BOS}
embedding $e_0$ with a uniformly sampled vocabulary token's embedding
before the forward pass. \emph{bi-replace-Lstar} leaves $e_0$ intact
and replaces $x_0^{(L^\star)}$ at the most-leaky layer $L^\star = 11$
with the mean of $x_k^{(L^\star)}$ across mid-sequence positions of
the same prompt. \emph{bi-replace-all} performs the latter at every
layer. We also measure the cross-prompt standard deviation of
$x_0^{(\ell)}$ at multiple layers across $8$ prompts.

\paragraph{Length-scaling and localization}
For length-scaling we capture logits at the first four layers and
heads of Llama-3.2-1B at sixteen sequence lengths and compare a
power-law fit to a piecewise model with breakpoint $N^\star$ by AIC.
The \emph{subspace ablation} projects out the linear direction in
$x_k^{(\ell^\star)}$ that best predicts $k$. The \emph{sink-head
ablation} zeroes the top-decile sink-reading heads against a
random-head control of equal size. The \emph{rank-1 MLP ablation}
replaces the FFN output at $(\ell^\star, k^\star = 64)$ by its
projection orthogonal to the dominant left-singular vector of the
per-prompt MLP outputs. All confidence intervals are $95\%$ bootstrap
intervals over $200$ prompt resamples.

\section{Results}
\label{sec:results}

\subsection{Sinks are token-anchored anchors, not content aggregators}

If sinks aggregate global content, the positions with the highest
attention magnitude should also be the positions from which content is
most decodable. They are not. Across all three models the
highest-attention positions have topic-probe accuracy at or below
chance, while the positions decodable above chance have sink
magnitudes three orders of magnitude smaller (Table~\ref{tab:exp-a}).

\begin{table}[h]
\centering
\small
\begin{tabular}{lcccc}
\toprule
& Model & Layer & $\mathcal{S}$ & $\mathcal{A}$ \\
\midrule
\multirow{3}{*}{Highest sink}
 & Qwen-2.5-3B & 5  & 0.991  & 0.219 \\
 & Qwen-2.5-3B & 20 & 0.963  & 0.219 \\
 & Qwen-2.5-3B & 25 & 0.946  & 0.219 \\
\midrule
\multirow{3}{*}{Highest content}
 & Mistral-7B  & 22 & 0.0038 & 0.415 \\
 & Mistral-7B  & 25 & 0.0034 & 0.402 \\
 & Mistral-7B  & 30 & 0.0063 & 0.395 \\
\bottomrule
\end{tabular}
\caption{Highest-attention and highest-content positions are disjoint.
Llama-3.2-1B follows the same pattern (appendix).}
\label{tab:exp-a}
\end{table}

Moving \texttt{BOS} to position $8$ transports the sink to position
$8$ in both Llama and Mistral, with attention magnitudes near
saturation (Table~\ref{tab:exp-b}). Replacing \texttt{BOS} at
position $0$ with a random uncommon token suppresses sink formation
there. The sink anchors to a token. Qwen, which does not auto-prepend
\texttt{BOS}, shows a sink at position $0$ with saturated magnitude as
a default-anchor fallback when no earlier token competes for the role.

\begin{table}[h]
\centering
\small
\begin{tabular}{llccc}
\toprule
Model & Variant & $k$ & $\mathcal{S}$ & $\mathcal{A}$ \\
\midrule
\multirow{3}{*}{Llama-3.2-1B}
 & default              & 0 & --             & 0.31 \\
 & \texttt{BOS} at $k=8$ & 8 & \textbf{0.92}  & 0.37 \\
 & \texttt{BOS} at $k=8$ & 0 & 0.59           & 0.31 \\
 & random at $k=0$       & 0 & $< 0.05$       & 0.34 \\
\midrule
\multirow{3}{*}{Mistral-7B-v0.3}
 & default              & 0 & 0.27           & 0.34 \\
 & \texttt{BOS} at $k=8$ & 8 & \textbf{0.94}  & 0.34 \\
 & random at $k=0$       & 0 & $< 0.40$       & 0.34 \\
\midrule
Qwen-2.5-3B
 & no \texttt{BOS}       & 0 & 1.00           & 0.36 \\
\bottomrule
\end{tabular}
\caption{The sink follows the \texttt{BOS} token across architectures.}
\label{tab:exp-b}
\end{table}

\subsection{Sink-reading heads carry sink-token identity}
\label{sec:results-cosine}

In Llama and Mistral, the top-decile heads by sink-reading strength
write output with markedly higher cosine consistency across prompts
than the bottom decile (Table~\ref{tab:cosine-consistency}). The
constant \texttt{BOS} token at position $0$ drives a constant signal
into the downstream heads. In Qwen, the top-decile heads write
input-varying output: the token at position $0$ varies across prompts.

\begin{table}[h]
\centering
\small
\resizebox{\columnwidth}{!}{%
\begin{tabular}{lcccc}
\toprule
Model & Heads & $\mathcal{C}$ top dec. & $\mathcal{C}$ bot dec. & $\mathrm{Sp}(\rho, \mathcal{C})$ \\
\midrule
Llama-3.2-1B   & 512  & $+0.66$ & $+0.14$ & $+0.48$ \\
Mistral-7B-v03 & 1024 & $+0.49$ & $+0.06$ & $+0.61$ \\
Qwen-2.5-3B    & 576  & $+0.40$ & $+0.49$ & $+0.06$ \\
\bottomrule
\end{tabular}
}
\caption{Per-head cosine consistency at $k = 64$, ranked by sink-reading
strength $\rho$.}
\label{tab:cosine-consistency}
\end{table}

The Qwen sink-token-identity probe confirms a single underlying
function (Table~\ref{tab:qwen-probe}). Three Qwen sink-reading heads
at layer $5$ give well-conditioned probe regressions on the sink-token
identity. L5/H10 and L5/H11 explain $91\%$ and $86\%$ of the variance
of the head's output projection from the sink-token identity alone,
with random-permutation null values near zero. Sink-reading heads
across all three architectures pass the identity of the token at
position $0$ to downstream attention; Llama and Mistral carry a
constant identity (\texttt{BOS}), Qwen carries a varying one. The
mechanism is the same. Section~\ref{sec:results-bosfp} grounds it
formally.

\begin{table}[h]
\centering
\small
\begin{tabular}{lcccc}
\toprule
Head & $\rho$ & PC1 var.\ fraction & $R^2_{\text{tok}}$ & $R^2_{\text{random}}$ \\
\midrule
L5/H10 & 1.000 & 0.605 & $+0.914$ & $-0.065$ \\
L5/H11 & 0.997 & 0.338 & $+0.860$ & $-0.057$ \\
L5/H15 & 0.979 & 0.347 & $+0.172$ & $-0.035$ \\
\bottomrule
\end{tabular}
\caption{Qwen sink-reading heads at layer 5 carry sink-token identity.
$R^2_{\text{tok}}$ is the cross-validated $R^2$ of the head's $W_O$
output projection on PC1 against a one-hot encoding of the sink-token
identity. Deeper-layer sink-reading heads produced ill-conditioned
probes and are excluded.}
\label{tab:qwen-probe}
\end{table}

\subsection{Leakage is distributed and structural}
\label{sec:results-distributed}

The saturated baseline reveals heads with statistically significant
absolute-position structure in every architecture
(Table~\ref{tab:leakage}). Peak per-head values are $0.05$ for Llama,
$0.11$ for Qwen, $0.24$ for Mistral; all of the top thirty heads per
model survive FDR correction at $q < 0.05$. The leakage is
content-independent: the mean $\Delta R^2$ on Llama is $0.012$ on
natural text and $0.011$–$0.013$ on random tokens.

\begin{table}[h]
\centering
\small
\resizebox{\columnwidth}{!}{%
\begin{tabular}{lcccc}
\toprule
Model & Tot heads & Peak $\Delta R^2$ & Med $\Delta R^2$ (top 30) & Sig.\ at $q<0.05$ \\
\midrule
Llama   & 512  & 0.049 & 0.029 & 30/30 \\
Qwen    & 576  & 0.108 & 0.041 & 30/30 \\
Mistral & 1024 & 0.241 & 0.047 & 30/30 \\
\bottomrule
\end{tabular}
}
\caption{Absolute-position leakage $\Delta R^2$ against the saturated
relative-offset baseline.}
\label{tab:leakage}
\end{table}

\subsection{Mechanism ablations expose two architectural components}
\label{sec:results-mechanism}

Three architectural ablations isolate the components
(Table~\ref{tab:mechanism}). Bidirectional ablation reduces $\Delta
R^2$ to between $9\%$ and $35\%$ of baseline; what survives
bidirectional is the residual-stream component
(Equation~\ref{eq:decomposition}). Identity-RoPE and scrambled-RoPE
increase $\Delta R^2$ by a factor of $2.1$ to $3.2$, making
position-varying residual-stream content directly readable from the
unrotated queries and keys.

\begin{table*}[h]
\centering
\small
\begin{tabular}{lcccc}
\toprule
Model & Baseline & Identity-RoPE & Scrambled-RoPE & Bidirectional \\
\midrule
Llama-3.2-3B    & 0.0143 [0.0137, 0.0149] & 0.0413 ($2.88\times$) & 0.0388 ($2.71\times$) & 0.0051 ($0.35\times$) \\
Qwen-2.5-3B     & 0.0153 [0.0144, 0.0162] & 0.0496 ($3.24\times$) & 0.0370 ($2.42\times$) & 0.0014 ($0.09\times$) \\
Mistral-7B-v0.3 & 0.0182 [0.0175, 0.0190] & 0.0391 ($2.14\times$) & 0.0386 ($2.12\times$) & 0.0022 ($0.12\times$) \\
\bottomrule
\end{tabular}
\caption{Mean $\Delta R^2$ across heads under three modifications of
the attention computation. Brackets are $95\%$ bootstrap CIs.}
\label{tab:mechanism}
\end{table*}

\subsection{Refined per-query binning}
\label{sec:results-binning}

Figure~\ref{fig:bins} shows mean $\Delta R^2$ by query position bin under causal and bidirectional attention for the three architectures, with the underlying values in Table~\ref{tab:bins}. At the first bin $[0, 5)$, where
leakage is overwhelmingly concentrated, the bidirectional value is
substantially smaller than the baseline in every architecture: the
ratio is $0.20$ for Qwen, $0.39$ for Llama, $0.51$ for Mistral. The
causal-mask share at this bin is $80\%$ in Qwen, $62\%$ in Llama, and
$49\%$ in Mistral.
\begin{table*}[h]
\centering
\small
\begin{tabular}{lccccccc}
\toprule
Bin in $i$ & $[0,5)$ & $[5,12)$ & $[12,25)$ & $[25,50)$ & $[50,100)$ & $[100,200)$ & $[200,256)$ \\
\midrule
Llama-3.2-3B baseline      & 0.296 & 0.080 & 0.032 & 0.016 & 0.011 & 0.008 & 0.003 \\
Llama-3.2-3B bidirectional & 0.114 & 0.090 & 0.036 & 0.018 & 0.010 & 0.005 & 0.001 \\
Mistral baseline           & 0.313 & 0.070 & 0.036 & 0.017 & 0.012 & 0.008 & 0.003 \\
Mistral bidirectional      & 0.161 & 0.078 & 0.033 & 0.016 & 0.008 & 0.002 & 0.001 \\
Qwen baseline              & 0.254 & 0.065 & 0.027 & 0.016 & 0.011 & 0.007 & 0.003 \\
Qwen bidirectional         & 0.050 & 0.015 & 0.005 & 0.003 & 0.002 & 0.001 & 0.001 \\
\bottomrule
\end{tabular}
\caption{Mean $\Delta R^2$ by query position bin under causal and
bidirectional attention.}
\label{tab:bins}
\end{table*}
Leakage drops by a factor of about $4\times$ from $[0, 5)$ to $[5, 12)$
and by another $2\times$ to $[12, 25)$. Beyond the first $25$ tokens
the contribution to head-averaged $\Delta R^2$ is small.
Equation~\ref{eq:partial-sum} predicts this position profile: $Z_i$
changes most rapidly at small $i$ and approaches its asymptote at large
$i$, so the causal-mask contribution to $\Delta R^2$ concentrates at
small $i$ by construction.

\subsection{Layer-wise depth profiles distinguish the three architectures}
\label{sec:results-depth}

The per-layer ratio $r_\ell$ exposes a third architectural distinction
(Table~\ref{tab:layerwise}). In Qwen, $r_\ell$ stays between $0.15$
and $0.30$ across all $36$ layers (mean $0.21$). In Llama-3.2-3B,
$r_\ell$ sits between $0.50$ and $0.70$ across all layers (mean
$0.62$), with no strong depth trend. In Mistral, $r_\ell$ grows from
about $0.50$ at layer $0$ to between $0.85$ and $0.90$ by layer $25$ of
$32$ (mean $0.66$). Sliding-window attention does not suppress
residual-stream propagation; in Mistral it allows the residual-stream
component to accumulate through depth.

\begin{table}[h]
\centering
\small
\begin{tabular}{lcc}
\toprule
Model & Mean $r_\ell$ & Depth trend \\
\midrule
Qwen-2.5-3B      & 0.21 & flat across 36 layers \\
Llama-3.2-3B     & 0.62 & flat across 28 layers \\
Mistral-7B-v0.3  & 0.66 & grows from $\approx 0.5$ at \\ & & $\ell = 0$ to $\approx 0.9$ at $\ell = 25$ \\
\bottomrule
\end{tabular}
\caption{Per-layer residual-stream share $r_\ell = \Delta
R^2_{\text{bidirectional}}(\ell) / \Delta R^2_{\text{baseline}}(\ell)$.}
\label{tab:layerwise}
\end{table}

\subsection{The residual-stream component traces to the position-0 trajectory}
\label{sec:results-bosfp}

The dynamical-system observation of Section~\ref{sec:mech} predicts
that $x_0^{(\ell)}$ at every layer is a fixed function of $e_0$ across
prompts. The prediction holds exactly in standard-RoPE architectures:
the cross-prompt standard deviation of $x_0^{(\ell)}$ at every layer
measured is exactly $0.0000$ in Llama-3.2-3B and Mistral-7B-v0.3, with
\texttt{BOS} fixed at position $0$ across the eight prompts tested
(Appendix~\ref{app:norms}). The standard deviation of $x_1^{(\ell)}$
at the same layer is on the order of $10^{-2}$ to $10^{-3}$, reflecting
variation in the second token across prompts. Sink-reading heads at
downstream positions therefore read a deterministic function of $e_0$.
In Qwen, $e_0$ varies with the input; the sink-token-identity
regression of Section~\ref{sec:results-cosine} is the direct readout of
the varying trajectory.

Three interventions on the position-$0$ trajectory test its
contribution to the residual-stream component
(Table~\ref{tab:bosfp}). Replacing the \texttt{BOS} embedding $e_0$
with a uniformly sampled vocabulary token's embedding reduces the
bidirectional first-bin $\Delta R^2$ from $0.115$ to $0.068$, a $40\%$
reduction. Replacing $x_0^{(L^\star)}$ at the most-leaky layer
$L^\star = 11$ with a prompt-specific mid-sequence mean reduces it to
$0.089$, a $22\%$ reduction. The \texttt{BOS} embedding alone
accounts for $40\%$ of the residual-stream component at early queries
in Llama-3.2-3B.

\begin{table}[h]
\centering
\small
\resizebox{\columnwidth}{!}{%
\begin{tabular}{lcc}
\toprule
Condition & First-bin $\Delta R^2$ & \% of bi.\ baseline \\
\midrule
Causal base                                       & 0.288 [0.279, 0.296] & 251\% \\
Bidirec base                                & 0.115 [0.108, 0.121] & 100\% \\
Bidirec + replace $e_0$                         & 0.068 [0.062, 0.075] & 60\% \\
Bidirec + replace $x_0^{(L^\star)}$             & 0.089 [0.082, 0.096] & 78\% \\
Bidirec + replace $x_0^{(\ell)}$ all $\ell$     & 0.147 [0.138, 0.154] & 128\% \\
\bottomrule
\end{tabular}
}
\caption{Position-$0$ interventions on first-bin $\Delta R^2$ in
Llama-3.2-3B. Brackets are $95\%$ bootstrap CIs. Replacing the
\texttt{BOS} embedding before the forward pass removes $40\%$ of the
residual-stream component at early queries.}
\label{tab:bosfp}
\end{table}

Replacing $x_0^{(\ell)}$ at every layer with a prompt-specific
mid-sequence mean does not reduce the residual-stream component; the
first-bin $\Delta R^2$ rises to $0.147$, $128\%$ of the bidirectional
baseline. The intervention introduces across-prompt variation at a
position that is normally constant, and the saturated regression reads
that variation as additional absolute-position structure. The
residual-stream component at early queries depends on $x_0$ being a
fixed fingerprint of $e_0$ across prompts, not on $x_0$ carrying
prompt-specific content.

The remaining $60\%$ that the embedding-replacement does not remove
comes from sources we do not isolate: layer-norm interactions with
position-correlated activations at mid-sequence positions, and
accumulated rotated writes through the network. The position-$0$
fingerprint accounts for a substantial fraction of the residual-stream
component at early queries; its remainder is open to future work.

\subsection{Length-scaling and targeted interventions}

Mean $\Delta R^2$ decays with sequence length.
A piecewise model that is flat below $N^\star = 80$ and decays as
$N^{-0.52}$ above is preferred over a single power law by AIC by
$\Delta\text{AIC} = -6.27$. The decay reflects the attenuation of the
causal-mask component as $Z_i$ approaches its limit.

\begin{figure*}[ht]
\centering
\includegraphics[width=0.8\textwidth]{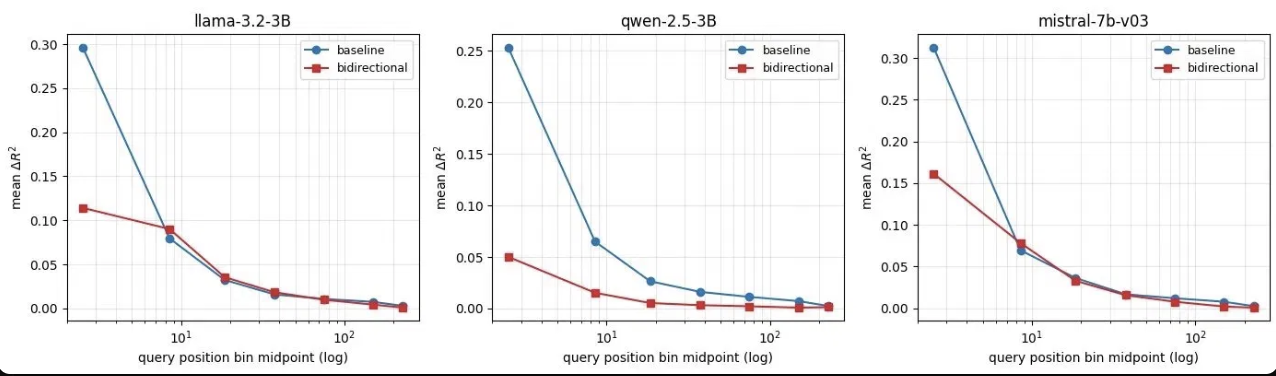}
\caption{Mean $\Delta R^2$ by query position bin under causal (baseline)
and bidirectional attention, for the three architectures. Query
position bin midpoints are plotted on a log scale.}
\label{fig:bins}
\end{figure*}

Targeted residual-stream interventions do not localize the leakage
(Table~\ref{tab:ablations}). Projecting out the linear direction that
best predicts absolute position drops mean $\Delta R^2$ by $3\%$.
Zeroing the top-decile sink-reading heads ($52$ heads) drops it by
$8\%$. A random-head control of equal size drops it by $17\%$. Both
targeted interventions are mechanically substantive and produce smaller
effects than a random ablation of the same size. The result is
consistent with one component living in the softmax normalization at
all heads and the other distributed across many residual-stream
directions.

\begin{table}[h]
\centering
\small
\resizebox{\columnwidth}{!}{%
\begin{tabular}{lccc}
\toprule
Condition & Heads zeroed & Mean $\Delta R^2$ & Reduc vs \ base \\
\midrule
Baseline                       & 0  & 0.0111 & --      \\
Rank-1 pos-probe subspace & 0  & 0.0107 & 3\%     \\
Top-decile sink-reading heads  & 52 & 0.0102 & 8\%     \\
Random-head control            & 52 & 0.0092 & 17\%    \\
\bottomrule
\end{tabular}
}
\caption{Targeted interventions on Llama-3.2-1B.}
\label{tab:ablations}
\end{table}

The rank-$1$ MLP ablation at $k^\star = 64$ removes up to $80\%$ of
the MLP output norm at the target but does not reduce topic-probe
accuracy in any of the nine layers tested
(Appendix~\ref{app:mlp-rank1}). Content production at content positions
is not concentrated in a single direction.

\subsection{Scale: composition shifts, magnitude is preserved}
\label{sec:results-scale}

In the Llama family the total leakage magnitude is approximately
constant across model sizes (Table~\ref{tab:scale}). The composition
shifts: the residual-stream share rises from $18\%$ at 1B to $35\%$
at 3B and $36\%$ at 8B. The causal-mask component is dominant at 1B
and roughly balanced with the residual-stream component at 3B and 8B.
The shift saturates between 3B and 8B at a fixed $\approx 64$/$36$
split.

\begin{table*}[h]
\centering
\small
\begin{tabular}{lcccc}
\toprule
Model & Baseline $\Delta R^2$ & Bidirectional $\Delta R^2$ & Residual-stream share & Causal-mask share \\
\midrule
Llama-3.2-1B  & 0.0121 & 0.0022 & 18\% & 82\% \\
Llama-3.2-3B  & 0.0143 & 0.0051 & 35\% & 64\% \\
Llama-3.1-8B  & 0.0135 & 0.0048 & 36\% & 64\% \\
\bottomrule
\end{tabular}
\caption{Scale composition shift. Total leakage magnitude is
approximately constant across 1B, 3B, 8B; the residual-stream share
of the leakage doubles from 1B to 3B and then saturates.}
\label{tab:scale}
\end{table*}

\section{Discussion}
\label{sec:discussion}

The leakage that RoPE-trained transformers carry in their attention
logits is the sum of two architectural components. The causal-mask
component is fixed by the per-query softmax denominator
(Equation~\ref{eq:partial-sum}) and concentrates at early query
positions where $Z_i$ is far from its asymptote. The residual-stream
component is distributed across many directions in the residual stream
and accumulates with depth in architectures that allow it. Both
components are present in all three architectures; their balance
differs.

The position-$0$ interventions ground the residual-stream component in
a concrete mechanism. Under causal attention $x_0^{(\ell)}$ is a
closed dynamical system from $e_0$, and the activation at every layer
is bit-identical across prompts when the token at position $0$ is
fixed. Replacing the \texttt{BOS} embedding alone reduces the
residual-stream component at early queries by $40\%$. The result
unifies two cross-architecture observations that previously looked
distinct. Sink-reading heads in Llama and Mistral write input-stable
output because they read $x_0^{(\ell)}$, which is constant across
prompts; sink-reading heads in Qwen carry sink-token identity because
they read the same $x_0^{(\ell)}$ trajectory, which varies with $e_0$
when $e_0$ is not the auto-prepended \texttt{BOS}. The same function
operates in all three architectures, on different inputs at position
$0$.

The architectural ordering is consistent across the measurements.
Qwen's NTK-scaled RoPE makes the residual-stream component the smallest
of the three (first-bin share $20\%$, layer-wise mean $r_\ell = 0.21$).
NTK scaling spreads the rotation frequencies further apart and reduces
the readability of residual-stream position content from the inner
product. Mistral's sliding-window attention does not suppress the
residual-stream component; $r_\ell$ rises from about $0.5$ at layer
$0$ to about $0.9$ at layer $25$. Standard RoPE in Llama sits between,
with a flat depth profile at $r_\ell \approx 0.6$.

The scale composition shift is informative. The Llama family
preserves total leakage magnitude across 1B, 3B, and 8B but
redistributes it from the causal-mask component (82\% at 1B) toward
the residual-stream component (36\% at 3B and 8B). The saturation
between 3B and 8B suggests that the architectural balance is set by
properties of the network that converge between $1$B and $3$B
parameters. The causal-mask component is determined by the geometry
of $Z_i$, which depends on the head-level distribution of $f$; the
residual-stream component depends on what the network learns to write
into $x_m^{(L)}$ and on the BOS-fingerprint trajectory at position
$0$. The convergence indicates that the second of these saturates at
moderate scales.

The negative localization results constrain the residual-stream
component. The rank-$1$ position-probe projection reduces downstream
leakage by only $3\%$. The MLP rank-$1$ ablation at content positions
removes up to $80\%$ of the FFN output norm without affecting decodable
content. The position-$0$ embedding intervention removes $40\%$ of the
first-bin residual-stream component, the largest single attribution we
can make; the other $60\%$ is distributed across sources the embedding
intervention does not touch. The empirical signature of the
residual-stream component is distribution across many directions and
many positions; whether this distribution is a learned optimization or
a consequence of the loss landscape under causal masking and RoPE is
open.

\section{Limitations}
\label{sec:limitations}

We measure leakage in attention logits and trace its
architectural sources, but do not connect the magnitude of either
component to a fully characterized downstream effect. The bidirectional ablation
removes the causal mask at inference time in networks whose weights
were trained under causal masking; whether the two-component
decomposition holds in a bidirectionally-trained model is open. The
position-$0$ embedding intervention removes $40\%$ of the residual-stream
component at early queries; the remaining $60\%$ comes from sources we
do not isolate (layer-norm interactions with mid-sequence
position-correlated activations, accumulated rotated writes through the
network), and characterizing these is a follow-up. 

\section{Conclusion}
\label{sec:conclusion}

The absolute-position structure that RoPE-trained transformers carry
in their attention logits is the sum of two architectural components.
The first is the per-query softmax normalization under causal masking,
which depends on the absolute query position through the cardinality
of the denominator. The second is position-varying structure in the
residual stream that RoPE rotation does not fully obscure, and traces
in part to a deterministic trajectory from the embedding of the token
at position $0$. Both components are present in all three
architectures, and the balance between them is set architecturally:
NTK scaling suppresses the residual-stream component, sliding-window
attention allows it to accumulate with depth, and standard RoPE in
Llama sits between. In the Llama family the composition shifts from
causal-mask-dominated at 1B to roughly balanced at 3B and 8B without
changing the total leakage magnitude. Attention sinks pass the
identity of the token at position $0$ to downstream attention through
a closed dynamical system; the architectural difference in
sink-reading head output stability reflects only the variation in
$e_0$ across architectures. Reconstruction of absolute position in a
RoPE-trained transformer is the joint operation of the causal mask and
the residual stream the network has learned to write.

\section{Acknowledgments}

Research carried out during Valeria Ruscio's postdoc at Sapienza University of Rome.

Research supported in part by European Union - Next Generation EU - namely by  the MUR-PRIN 2022 project "2022REWNTE - Artificial Intelligence  algorithms to track and detect Covid-19 vaccine-related infodemic on  social media" - CUP no.B53D23020690006; in part by the projects FAIR under Grant PE0000013 and SERICS under Grant PE00000014 under the MUR National Recovery and Resilience Plan funded by the European Union-NextGenerationEU, and in part by the project Neural Reasoning over Open Data (NEREO) funded by the Italian Ministry of Education and Research (PRIN) under Grant 2022AEFHA, and in part by the project SEED funded by Sapienza University of Rome.

\bibliography{custom}

\newpage

\appendix
\section{Position-0 trajectory: norm measurements}
\label{app:norms}

We measure the norm and cross-prompt standard deviation of the
residual-stream activation at position $0$ at multiple layers of
Llama-3.2-3B and Mistral-7B-v0.3, using eight prompts in which the
token at position $0$ is the auto-prepended \texttt{BOS} (\texttt{id}
$= 128000$ for Llama, $1$ for Mistral) and subsequent tokens vary.

\begin{table}[h]
\centering
\small
\begin{tabular}{lccc}
\toprule
Layer & Mean $\|x_0^{(\ell)}\|$ & Std across prompts \\
\midrule
\multicolumn{3}{l}{\textit{Llama-3.2-3B}} \\
embedding   & $1.13$   & $0.00$ \\
$\ell = 0$  & $19.92$  & $0.00$ \\
$\ell = 5$  & $593.68$ & $0.00$ \\
$\ell = 11$ & $593.29$ & $0.00$ \\
$\ell = 15$ & $594.70$ & $0.00$ \\
$\ell = 20$ & $594.55$ & $0.00$ \\
$\ell = 27$ & $248.03$ & $0.00$ \\
\midrule
\multicolumn{3}{l}{\textit{Mistral-7B-v0.3}} \\
embedding   & $0.14$   & $0.00$ \\
$\ell = 0$  & $6.02$   & $0.00$ \\
$\ell = 5$  & $265.00$ & $0.00$ \\
$\ell = 11$ & $265.07$ & $0.00$ \\
$\ell = 15$ & $265.10$ & $0.00$ \\
$\ell = 20$ & $268.78$ & $0.00$ \\
$\ell = 31$ & $325.84$ & $0.00$ \\
\bottomrule
\end{tabular}
\caption{Norm of $x_0^{(\ell)}$ at multiple layers across eight prompts.
The cross-prompt standard deviation is exactly zero at every layer of
both architectures: the position-$0$ trajectory is fully determined by
the \texttt{BOS} embedding. For reference, the cross-prompt variance
of $x_1^{(\ell)}$ at $\ell = 11$ is $1.24 \times 10^{-2}$ in Llama and
$1.26 \times 10^{-3}$ in Mistral, reflecting the variation in the
second token.}
\label{tab:norms}
\end{table}

The peak mid-sequence norm of $x_0^{(\ell)}$ is $593$ in Llama-3.2-3B
and $265$ in Mistral-7B. In both architectures the norm at position
$0$ is more than an order of magnitude larger than the typical
mid-sequence norm at downstream positions (around $14\times$ for both
models when measured at the same layer). The position-$0$ activation
acts as an anomalously high-magnitude fixed vector that downstream
attention reads.

\section{Rank-1 MLP ablation: per-layer values}
\label{app:mlp-rank1}

\begin{table}[h]
\centering
\small
\begin{tabular}{lccc}
\toprule
Model & L & $\Delta \mathcal{A}$ & Frac MLP norm removed \\
\midrule
\multirow{3}{*}{Llama-3.2-1B}
 & 15 & $-0.015$ & 0.80 \\
 & 14 & $-0.015$ & 0.63 \\
 & 13 & $-0.004$ & 0.43 \\
\midrule
\multirow{3}{*}{Qwen-2.5-3B}
 & 35 & $+0.027$ & 0.17 \\
 & 34 & $+0.011$ & 0.37 \\
 & 33 & $\phantom{+}0.000$ & 0.34 \\
\midrule
\multirow{3}{*}{Mistral-7B-v03}
 & 31 & $-0.020$ & 0.66 \\
 & 30 & $+0.012$ & 0.30 \\
 & 29 & $+0.011$ & 0.17 \\
\bottomrule
\end{tabular}
\caption{Rank-1 MLP ablation at $k^\star = 64$ removes the majority of
the MLP output norm at the target but does not suppress decodable
content.}
\label{tab:rank1-app}
\end{table}

\section{Per-query-position profile under bidirectional attention}
\label{app:bins-bidirectional}

Section~\ref{sec:results-binning} presents the per-query-position
$\Delta R^2$ profile under causal attention and shows that leakage
concentrates at small $i$. The same binning under bidirectional
attention is the cleanest direct test of the causal-mask pathway: if
the causal mask is the source of the position dependence at small $i$,
the bidirectional curve should be flat in $i$ relative to the steep
decay of the causal curve.

Figure~\ref{fig:bins} (main text) and Table~\ref{tab:bins} give the
underlying numbers. The bidirectional first-bin $\Delta R^2$ falls
sharply relative to the causal first-bin value in every architecture:
the causal-to-bidirectional gap at $[0, 5)$ is $0.182$ in Llama-3.2-3B,
$0.152$ in Mistral, and $0.204$ in Qwen. The gap closes by $[12, 25)$
in all three architectures and the two curves become indistinguishable
beyond $[25, 50)$.

The bidirectional curve is itself not perfectly flat at small $i$. It
decays from $0.114$ to $0.090$ between $[0, 5)$ and $[5, 12)$ in
Llama-3.2-3B, and from $0.161$ to $0.078$ between the same two bins in
Mistral. Two effects contribute. The position-$0$ trajectory has
anomalously high norm (Appendix~\ref{app:norms}) and influences
attention weights at early queries more than at later ones, even with
the causal mask removed. The residual-stream component itself has
position dependence that is not captured by causal-mask geometry alone.
The decay of the bidirectional curve at small $i$ is the empirical
signature of this position-dependent residual-stream content.

The Qwen profile is the cleanest. The bidirectional curve sits an
order of magnitude below the causal curve at $[0, 5)$ ($0.050$ vs
$0.254$) and decays smoothly with $i$ without crossing the causal
curve at any bin. NTK scaling suppresses the residual-stream
contribution at all positions, leaving the causal-mask geometry as the
dominant source of the position-dependent attention weights at early
queries.

In Llama-3.2-3B and Mistral the bidirectional curve briefly exceeds the
causal curve in the $[5, 12)$ and $[12, 25)$ bins (Mistral) or just
$[5, 12)$ (Llama). The crossing is at $\Delta R^2$ values below $0.10$
and the two curves are statistically indistinguishable within their
bootstrap confidence intervals over that range. The crossing reflects
the larger absolute residual-stream contribution in standard-RoPE
architectures, which becomes visible once the dominant causal-mask
contribution at small $i$ is removed.

\end{document}